\documentclass[letterpaper, 10 pt, conference]{ieeeconf}

\usepackage[latin1]{inputenc}
\usepackage{tikz}
\usetikzlibrary{patterns}
\usepackage{verbatim}
  
\usepackage{amsthm}
\usepackage{amsfonts}

\usepackage{graphicx}
\usepackage{enumerate}
\usepackage{color}
\usepackage{bm}
\usepackage{url}
\usepackage{mathtools}

\usepackage{setspace} 
\IEEEoverridecommandlockouts 
\title{\huge
Restricted Discrete Invariance and Self-Synchronization For Stable Walking of Bipedal Robots
}
\author{Hamed Razavi$^{1}$, Anthony M. Bloch$^{2}$, Christine Chevallereau$^3$ and J. W. Grizzle$^4$
\thanks{$^1$Hamed Razavi and $^2$Anthony Bloch are with the Mathematics Department of the University of Michigan, Ann Arbor, MI, USA,
        {\tt \{razavi, abloch\}@umich.edu}}%
\thanks{$^3$ Christine Chevallereau is with LUNAM Universit\'e, IRCCyN, CNRS, France,
{\tt Christine.Chevallereau}@\tt{irccyn.ec-nantes.fr}}
\thanks{$^{4}$ J. W. Grizzle is with the Electrical Engineering and Computer Science Department of the University of Michigan, Ann Arbor, MI, USA,
        {\tt grizzle@umich.edu}}%
}

\begin{document}
\maketitle
\thispagestyle{empty}
\pagestyle{empty}
\newcommand{\nvar}[2]{%
    \newlength{#1}
    \setlength{#1}{#2}
}

\nvar{\dg}{0.3cm}
\def\dw{0.25}\def\dh{0.5}
\nvar{\ddx}{1cm}

\def\link{\draw [double distance=0.25mm, very thick] (0,0)--}
\def\joint{%
    \filldraw [fill=white] (0,0) circle (8pt) ;
        \filldraw[fill=black] (0,0) circle (8pt);
}

\def\robotbase{%
    \draw (-2.5,-.1)-- (1.5,-.1);
    \fill[pattern=north east lines] (-2.5,-.1) rectangle (1.5,-.5);
}

\def\robotbasetwo{%
    \draw (-3.5,-.1)-- (.5,-.1);
    \fill[pattern=north east lines] (-3.5,-.1) rectangle (.5,-.5);
}
\def\robotbasethree{%
    \draw (-3.8,0)-- (.5,0);
    \fill[pattern=north east lines] (-3.8,0) rectangle (.5,-.5);
}

\def\robotbasefour{%
    \draw[rounded corners=8pt] (-\dw,-\dh)-- (-\dw, 0) --
        (0,\dh)--(\dw,0)--(\dw,-\dh);
    \draw (-0.5,-\dh)-- (0.5,-\dh);
    \fill[pattern=north east lines] (-0.5,-.75) rectangle (0.5,-\dh);
}
\def\Lone{3.4*.85}
\def\thone{70}
\def\thtwo{50}
\def\ththree{\thone+5}
\def\thfour{15}
\def\ArcRadius{\Lone}
\def\ArcAngle{\thone}

\vspace{4cm}
\def\linktwo{\draw [double distance=1.5mm, very thick] (0,0)--}
\def\jointtwo{%
    \filldraw [fill=white] (0,0) circle (5pt);
    \fill[black] circle (2pt);
}
{\def\jointone{%
    \filldraw [fill=white] (0,0) circle (3pt);
   }
{\def\jointthree{%
    \filldraw [fill=white] (-.1,-.1)--(-.1,.1)--(.1,.1)--(.1,-.1)--(-.1,-.1);
}



\def\grip{%
    \draw[ultra thick](0cm,\dg)--(0cm,-\dg);
    \fill (0cm, 0.5\dg)+(0cm,1.5pt) -- +(0.6\dg,0cm) -- +(0pt,-1.5pt);
    \fill (0cm, -0.5\dg)+(0cm,1.5pt) -- +(0.6\dg,0cm) -- +(0pt,-1.5pt);
}

\newcommand{\angann}[2]{%
    \begin{scope}[red]
    \draw [dashed, red] (0,0) -- (1.2\ddx,0pt);
    \draw [-] (\ddx,0pt) arc (0:#1:\ddx);
    \node at (#1/2-2:\ddx+8pt) {#2};
    \end{scope}
}

\newcommand{\lineanntwo}[4][0.5]{%
    \begin{scope}[rotate=#2, blue,inner sep=2pt]
        \draw[dashed, blue!40] (0,0) -- +(0,#1)
            node [coordinate, near end] (a) {};
        \draw[dashed, blue!40] (#3,0) -- +(0,#1)
            node [coordinate, near end] (b) {};
        \draw[|<->|] (a) -- node[fill=white] {#4} (b);
    \end{scope}
}

\def\thetaone{30}
\def\thetatwo{30}
\def\Ltwo{2}
\def\thetathree{30}
\def\Lthree{1}

\begin{abstract}
Models of bipedal locomotion are hybrid, with a continuous component often generated by a Lagrangian plus actuators, and a discrete component where leg transfer takes place. The discrete component typically consists of a locally embedded co-dimension one submanifold in the continuous state space of the robot, called the switching surface,  and a reset map that provides a new initial condition when a solution of the continuous component intersects the switching surface.  The aim of this paper is to identify a low-dimensional submanifold of the switching surface, which, when it can be rendered invariant by the closed-loop dynamics, leads to asymptotically stable periodic gaits. The paper begins this process by studying the well-known 3D Linear Inverted Pendulum (LIP) model, where analytical results are much easier to obtain. A key contribution here is the notion of \textit{self-synchronization}, which refers to the periods of the pendular motions in the sagittal and frontal planes tending to a common period. The notion of invariance resulting from the study of the 3D LIP model is then extended to a 9-DOF 3D biped. A numerical study is performed to illustrate that asymptotically stable walking may be obtained.
\end{abstract}

\section{Introduction}
While steady progress is being made on the design and analysis of control algorithms for achieving asymptotically stable walking in high-dimensional 3D bipedal robots, the problem remains a very active research area.  The control method most widely used on humanoid robots is based on the Zero Moment Point (ZMP) criterion \cite{KaMaHaKaKaFuHi03,VUBOPO06,KaKaKaFuHaYoHi06,GO90,GOS99}, which imposes restrictions on the gaits, such as walking flat-footed. A foot positioning method based on ``capture points'' has been introduced in Pratt et. al \cite{pratt2012capturability}, and allows some gaits with partial foot contact with the ground.  Both of these methods are based on the Linear Inverted Pendulum model (LIP) \cite{kajita2001}. Ames \cite{ames2007geometric} and Greg et. al \cite{gregg2008reduction} developed Routhian reduction for fully-actuated 3D bipeds to allow some controllers developed for planar robots an immediate extension to the 3D setting. Grizzle et. al \cite{westervelt2007feedback,grizzle2014atrias}, use virtual constraints on an underactuated robot to create an invariant submanifold in the closed-loop hybrid model, and reduce the design of asymptotically stable motions to the study of a low-dimensional system, the hybrid zero dynamics;  initial 3D experiments are reported in \cite{busspreliminary}. These last two methods do not rely on simplified models.

The aim of the present paper is to identify a low-dimensional submanifold of the switching surface--instead of the entire hybrid model--which, when it can be rendered invariant by the closed-loop dynamics, leads to asymptotically stable periodic gaits. The paper begins this process by studying the well-known 3D LIP model, where analytical results are much easier to obtain. A key contribution here is the notion of \textit{self-synchronization}, which refers to the periods of the pendular motions in the sagittal and frontal planes tending to a common period. The notion of invariance resulting from the study of the 3D LIP model is then extended to a 9-DOF 3D biped, corresponding to a simplified version of the ATRIAS robot designed by Hurst \cite{GRIMES2012, grizzle2014atrias, hubicki2014atrias}. A numerical study is performed to illustrate that asymptotically stable walking may be obtained.

The  paper is organized as follows. In Section \ref{Self_Synchronization_3DLIP} we introduce a discrete invariant gait for the 3D LIP. Then the concept of synchronization is introduced and it is proven that under the discrete invariant gait the 3D LIP will have a periodic motion in which the oscillations in the sagittal and frontal planes are self-synchronized. In Section \ref{9-DOF_3D_Biped}, after introducing the model of the  9-DOF 3D biped, inspired by the 3D LIP, we define a discrete invariant gait for the  9-DOF 3D biped. Then we perform a reduction based on this gait, and corresponding to this reduction we define a restricted Poincar\'e map. After presenting an example of the controllers which enforce the discrete invariant gait, we show that the eigenvalues of the restricted poincare map lie in the unit circle and the walking motion of the biped is stable. The final section includes the concluding remarks.

\section{Discrete Invariance and Self-Synchronization of the 3D LIP}
\label{Self_Synchronization_3DLIP}
In this section we introduce the notion of discrete invariant gait for the 3D LIP and we will show that under this gait the oscillations in the sagittal plane and frontal plane can be self-synchronized.
The 3D LIP is known to be a simple walking model that can capture many properties of more complex 3D walking models. For example, in \cite{koolen2012capturability} the authors introduce the concept of \textit{capturability} and find what they call \textit{capture regions} based on different types of the 3D LIP model. Later they apply these results to a 12-DOF 3D biped \cite{pratt2012capturability}. To generate walking patterns for a humanoid robot, Kajita et. al \cite{kajita2001} use the 3D LIP model. Even though the 3D LIP models are simple and have been found to be successful, they have some limitations. For example, the energy loss due to the impact is not usually considered in these models and some researchers have modified the LIP model to overcome this limitation \cite{grizzle2014models}. 
\begin{figure}
\begin{center}
\includegraphics[scale=.27]{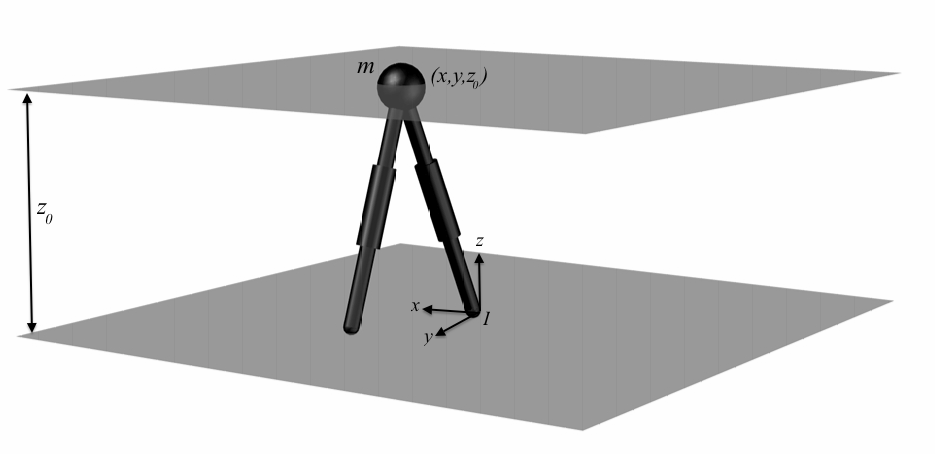}
\end{center}
\caption{3D LIP Biped Model}
\label{3d_lip_fig}
\end{figure}
We first define a hybrid model of a 3D LIP biped with massless legs as shown in Figure \ref{3d_lip_fig}. Let $(x, y, z)$ denote the position of the point mass, $m$, of the 3D LIP in the inertial coordinate frame, $I$, centered at the point of support. Assuming that the mass moves in the plane $z = z_0$, $(x, y, \dot{x}, \dot{y})$ defines a set of generalized coordinates for the 3D LIP. Without loss of generality we can assume $m = 1$. The equations of motion in the continuous phase are \cite{kajita2001}:
\begin{eqnarray*}
\begin{array}{c}
\ddot{x} = \omega^2 x\\
\ddot{y} = \omega^2 y,
\end{array}
\end{eqnarray*}
where $\omega = \sqrt{g/z_0}$. Also, we define the discrete map as follows
\begin{eqnarray}
\begin{array}{llc}
\dot{x}^+ &=& \dot{x}^-\\
\dot{y}^+ &=& -\dot{y}^-\\
x^+ &=& x_{FH}^-\\
y^+ &=& -y_{FH}^-
\end{array}
 (x^-, y^-, \dot{x}^-, \dot{y}^-) \in \mathcal{S},
\label{impact_map_3dlip}
\end{eqnarray}
where $(x_{FH}^-, y_{FH}^-)$ is the position vector from the swing leg end to the point mass represented in $I$ right before the impact and $\mathcal{S}$ is the switching manifold which is defined as
\begin{equation*}
\mathcal{S} = \{ (x,y, \dot{x}, \dot{y}) | f(x,y) = 0\},
\end{equation*}
for a smooth function $f: \mathcal{S}\rightarrow \mathbb{R}$ with rank 1.  

The reason that in equation (\ref{impact_map_3dlip}) sign of $\dot{y}$ changes after the impact and $y^+ = -y_{FH}^-$ (and not $y_{FH}^-$) is that by changing the support point we switch the direction of the $y$ coordinate of the inertial frame which is centered at the new support point. 

\theoremstyle{definition}
\newtheorem{3d_lip_invariant_gait}{Definition}
\begin{3d_lip_invariant_gait}
(3D LIP Discrete Invariant Gait) Consider a 3D LIP as described in Figure \ref{3d_lip_fig}. Suppose that at the beginning of the step $(x, y) = (-x_0, y_0)$ for $x_0>0$ and $y_0>0$.  We say that the 3D LIP takes an $(x_0,y_0)-$invariant step if the switching manifold is defined as
\begin{equation}
\mathcal{S} = \{ (x,y, \dot{x}, \dot{y}) | x^2+y^2 = x_0^2+y_0^2\},
\label{switching_manifold_3dlip}
\end{equation}
and at the moment of double support $(x_{FH}, y_{FH}) = (-x_0,-y_0)$. 
\end{3d_lip_invariant_gait}

\theoremstyle{definition}
\newtheorem{why_invariance}[3d_lip_invariant_gait]{Remark}
\begin{why_invariance}
This gait is called $(x_0, y_0)$-invariant because we are assuming that the swing leg is exactly at the desired position so that $(x_{FH}^-, y_{FH}^-) = (-x_0, -y_0)$ at impact and hence we are assuming that at the beginning of the current step and the next step $(x,y) = (-x_0, y_0)$. We refer to this as \textit{Discrete Invariance} since it only imposes conditions on the discrete phase of motion.  
\end{why_invariance}

\theoremstyle{definition}
\newtheorem{orbital_energies}[3d_lip_invariant_gait]{Definition}
\begin{orbital_energies}
In the 3D LIP, $E_x = \dot{x}^2-\omega^2 x^2$ and $E_y = \dot{y}^2-\omega^2y^2$ are called the orbital energies in the $x$ and $y$ directions \cite{kajita2001}. 
\label{orbital_energies}
\end{orbital_energies}

\theoremstyle{plain}
\newtheorem{constant_kinetic_energy}[3d_lip_invariant_gait]{Proposition}
\begin{constant_kinetic_energy}
Let $K_0$ be the kinetic energy of the 3D LIP at the beginning (end) of the step. Suppose that the 3D LIP biped completes an $(x_0, y_0)$-invariant step and $K_1$ is the kinetic energy of the 3D LIP at the beginning (end) of the next step. We have $K_1 = K_0$. 
\begin{proof}
If $E_{x_0}$ and $E_{y_0}$ are the orbital energies at the beginning of the step,
\begin{eqnarray}
\nonumber
E_{x_0}+E_{y_0} &=& \dot{x}_0^2+\dot{y}_0^2-\omega^2 (x_0^2+y_0^2)\\
\label{Ex0_plus_Ey0}
&=& 2K_0-\omega^2 r_0^2,
\end{eqnarray}
where $K_0$ is the kinetic energy at the beginning of the step and $r_0^2 = x_0^2+y_0^2$. 
On the other hand, if right before the impact $x = x_1$, $y = y_1$, $E_x = E_{x_1}^-$, and $E_y = E_{y_1}^-$, then\begin{eqnarray*}
E_{x_1}^-+E_{y_1}^- &=& 2K_1^--\omega^2 r_1^2,
\end{eqnarray*}
where $K_1^-$ is the kinetic energy right before the impact and $r^2_1 = x^2_1+y^2_1$. However, because by definition of the switching surface at impact $x^2_1+y^2_1=r_0^2$, from the equation above
\begin{eqnarray}
E_{x_1}^-+E_{y_1}^- &=& 2K_1^--\omega^2 r_0^2.
\label{Ex1_plus_Ey1}
\end{eqnarray}
Since, the orbital energies are conserved quantities during each step, $E_{x_0} = E_{x_1}^-$ and $E_{y_0}= E_{y_1}^-$. Therefore, comparing equations (\ref{Ex0_plus_Ey0})  and (\ref{Ex1_plus_Ey1}), we have $K_1^- = K_0$. By equation (\ref{impact_map_3dlip}) there is no loss in velocities due to impact, therefore, $K_1 = K_0$, i.e. the kinetic energy at the beginning of the next step is equal to the kinetic energy at the beginning of the current step. 
\end{proof}
\label{constant_kinetic_energy}
\end{constant_kinetic_energy}

\theoremstyle{definition}
\newtheorem{self_synch_def}[3d_lip_invariant_gait]{Definition}
\begin{self_synch_def}
(Synchronization) Suppose that at the beginning of a step $E_x>0$ and $E_y<0$. The solution in the step is said to be \textit{synchronized} if  $\dot{y} = 0$ when $x = 0$. 
\label{self_synch_def}
\end{self_synch_def}
One can easily check that if at the beginning of the step $(x, y) = (-x_0,y_0)$ and the solution in the step is synchronized, then at the end of the step $(x, y) = (x_0,y_0)$.
\theoremstyle{plain}
\newtheorem{periodicity_3dlip}[3d_lip_invariant_gait]{Proposition}
\begin{periodicity_3dlip}
If the initial conditions are such that the first step of the 3D LIP under the $(x_0, y_0)$-invariant step is synchronized then the subsequent step will also be synchronized and the 3D LIP follows a 1-periodic motion.  
\label{periodicity_3dlip}
\end{periodicity_3dlip}
\begin{proof}
The proof follows from the fact that in a synchronized solution the velocities at the end of the step are equal to the velocities at the beginning of the step except that the velocity in the $y$ direction is reversed. 
\end{proof}
\theoremstyle{definition}
\newtheorem{sync_measure}[3d_lip_invariant_gait]{Definition}
\begin{sync_measure}
Suppose that for the 3D LIP at the beginning of the step $x = -x_0<0$, $y = y_0>0$ and the initial velocities are $\dot{x}_0>0$ and $\dot{y}_0 <0$. The \textit{synchronization measure} for this step is defined as $L_0 = \dot{x}_0\dot{y_0}+\omega^2 x_0 y_0$.
\label{sync_measure}
\end{sync_measure}
In the next proposition we see that the solution in the step is synchronized if and only if the synchronization measure is zero. Later, to show that the motion is self-synchronized we show that the synchronization measure approaches zero in the subsequent steps. 
\theoremstyle{plain}
\newtheorem{cond_for_self_synch}[3d_lip_invariant_gait]{Proposition}
\begin{cond_for_self_synch}
Suppose that the 3D LIP has an $(x_0, y_0)$-invariant gait with initial velocities $\dot{x} = \dot{x}_0>0$ and $\dot{y} = \dot{y}_0<0$.  Also, suppose that $E_{x}>0$ and $E_{y}<0$. Let $L_0$ denote the synchronization measure of this step. The motion of the 3D LIP is 1-periodic if and only if $L_0 = 0$. 
\label{cond_for_self_synch}
\end{cond_for_self_synch}
\begin{proof}
With the given initial conditions the solution to the 3D LIP system in the continuous phase is
\begin{eqnarray}
\label{x3dlip}
x(t)&=&-x_0\cosh(\omega t)+\frac{\dot{x}_0}{\omega} \sinh(\omega t)\\
\label{y3dlip}
y(t)& =&y_0\cosh(\omega t)+\frac{\dot{y}_0}{\omega} \sinh(\omega t).
\end{eqnarray}
We want to find $(\dot{x}_0, \dot{y}_0)$ such that the solution that starts from $(-x_0, y_0)$ is synchronized. Therefore, if we set the second equation above to zero, we find the time, $t_y$, which it takes for $\dot{y}$ to become zero:
\begin{eqnarray}
\label{ty}
\tanh(\omega t_y)=-\frac{\dot{y}_0}{y_0\omega}
\end{eqnarray}
Similarly, from equation (\ref{x3dlip}) the time $t_x$, at which $x=0$ is found by the following equation:
\begin{eqnarray}
\label{tx}
\tanh(\omega t_x)=\frac{x_0 \omega}{\dot{x}_0}
\end{eqnarray}
According to the definition of the synchronization, synchronization occurs if $t_x = t_y$, therefore from equations (\ref{ty}) and (\ref{tx}), the motion of the 3D LIP in this step is synchronized if and only if
\begin{eqnarray*}
-\frac{\dot{y}_0}{y_0\omega} = \frac{x_0 \omega}{\dot{x}_0}
\end{eqnarray*}
Hence, from this equation we have synchronization if and only if $\dot{x}_0\dot{y}_0+\omega^2 x_0 y_0 = 0$. This result together with Proposition \ref{periodicity_3dlip} complete the proof. 
\end{proof}

So far, we have shown that if the initial conditions are such that $L_0 = 0$, the solution is synchronized. We are also interested in stability in the synchronization. Interestingly, as we will show in the next proposition, the discrete invariant gait can result in self-synchronization of the 3D LIP, that is, eventually the synchronization measure approaches zero. 
\theoremstyle{plain}
\newtheorem{self_sync_stability}[3d_lip_invariant_gait]{Proposition}
\begin{self_sync_stability}
Suppose that the 3D LIP biped model takes an $(x_0, y_0)$-invariant step with initial velocities $\dot{x} = \dot{x}_0>0$ and $\dot{y} = \dot{y}_0<0$. Suppose that $K_0$ is the initial kinetic energy of the system and $K_0 - \omega^2 x_0 y_0>0$. Let $L_0$ and $L_1$ denote the synchronization measure in the current step and next step, respectively. 
We have
$$\lim_{L_0\rightarrow 0} \frac{L_{1}}{L_0} = -\lambda.$$
where
\begin{eqnarray}
\lambda = 1 - \frac{2\omega^2 (y_0^2 - x_0^2)}{\omega^2 (y_0^2 - x_0^2)+2\sqrt{K_0^2-\omega^4x_0^2 y_0^2}}.
\label{lambda_eqn}
\end{eqnarray}
If $y_0 > x_0$, we have $|\lambda|<1$.
\begin{proof}
One can easily check that the orbital energies, $E_x$, $E_y$, and synchronization measure, $L = \dot{x}\dot{y}-\omega^2 xy$  are conserved quantities in the continuous phase of the motion. Also, at impact $x^2+y^2 = x_0^2+y_0^2$. Therefore, if the state right before the impact is $(x_1, y_1, \dot{x}_1^-,\dot{y}^-_1)$,
\begin{eqnarray}
\label{at_impact1}
(\dot{x}_1^-)^2-\omega^2 x_1^2 &=& \dot{x}_0^2 - \omega^2 x_0^2\\
(\dot{y}_1^-)^2-\omega^2 y_1^2 &=& \dot{y}_0^2 - \omega^2 y_0^2\\
\dot{x}_1^- \dot{y}_1^- - \omega^2 x_1 y_1 &=& \dot{x}_0 \dot{y}_0 +\omega^2 x_0 y_0\\
 x_1^2 + y_1^2 &=& x_0^2 + y_0^2.
 \label{at_impact4}
\end{eqnarray}
By Proposition \ref{cond_for_self_synch}, if $L_0 = 0$ the motion is periodic and it can be checked that on this periodic orbit we have the following equations for the orbital energies $E_x^*$ and $E_y^*$:
\begin{eqnarray}
\label{E_x_star}
E_x^* = K_0+\sqrt{K_0^2-\omega^4 x_0^2 y_0^2}-\omega^2 x_0^2\\
E_y^* = K_0 -\sqrt{K_0^2 - \omega^4 x_0^2 y_0^2}-\omega^2 y_0^2.
\label{E_y_star}
\end{eqnarray}
If $L_0 = 0$ then $x_1 = x_0$, $y_1 = y_0$, $\dot{x}_1^- = \dot{x}_0$ and $\dot{y}_1^- = -\dot{y}_0$. Now, assuming $L_0$ is infinitesimally small, we have
\begin{eqnarray*}
x_1 = x_0 + \delta x_0,& y_1 = y_0 + \delta y_0\\
\dot{x}_1^- = \dot{x}_0 + \delta \dot{x}_0,& \dot{y}_1^- = -\dot{y}_0 + \delta \dot{y}_0.
\end{eqnarray*}
If we substitute these equations into equations (\ref{at_impact1}-\ref{at_impact4}), we can show that, 
\begin{eqnarray}
\lim_{L_0 \rightarrow 0} \frac{\delta x_0}{L_0} = \frac{2 y_0}{E_{y*}-E_{x*}}.
\label{formula_for_deltax}
\end{eqnarray}
By definition, if $\dot{x}_1$ and $\dot{y}_1$ are the velocities at the beginning of the next step, $L_1 = \dot{x}_1\dot{y}_1+ \omega^2 x_0 y_0$. Also, because $\dot{x}\dot{y} - \omega^2 x y$ is conserved during the step, and $(\dot{x}_1,\dot{y}_1) = (\dot{x}_1^-, -\dot{y}_1^-)$, we have $L_0 =- \dot{x}_1\dot{y}_1-\omega^2 x_1 y_1$. 
From these last two equations for $L_0$ and $L_1$, and equation (\ref{formula_for_deltax}) we can show that
\begin{eqnarray}
\lim_{L_0\rightarrow 0}\frac{L_{1}}{L_0} = -1-\frac{2\omega^2(y_0^2-x_0^2)}{E_y^*-E_x^*}. 
\label{lim_L0_L1}
\end{eqnarray}
From equations (\ref{E_x_star}) and (\ref{E_y_star}), 
\begin{eqnarray*}
E_y^*-E_x^* = -2\sqrt{K_0^2-\omega^4 x_0^2 y_0^2}-\omega^2(y_0^2-x_0^2),
\end{eqnarray*}
Therefore, substituting this into equation (\ref{lim_L0_L1}) we get equation (\ref{lambda_eqn}) for $\lambda$. From equation (\ref{lambda_eqn}), if $y_0>x_0$ we have $|\lambda|<1$. 
\label{self_sync_stability}
\end{proof}
\end{self_sync_stability}
As we will see in the analysis of the Poincar\'e map, this proposition proves that if $y_0>x_0$ and $K_0^2-\omega^4 x_0^2 y_0^2>0$, the motion of the 3D LIP is self-synchronized. 
Inspired by this proposition we define an alternative generalized coordinate for the 3D LIP which allows us to simplify the analysis of stability. 
\theoremstyle{definition}
\newtheorem{alternative_coor_3dlip}[3d_lip_invariant_gait]{Definition}
\begin{alternative_coor_3dlip}
For the 3D LIP define $\alpha = \tan^{-1} (\frac{x}{y})$, $r = \sqrt{x^2+y^2}$, $\gamma = \dot{x} \dot{y}$ and $v= \sqrt{\dot{x}^2+\dot{y}^2}$. Then, if $y\ne 0$, $(r, \alpha, \gamma, v)$ defines a coordinate system for the 3D LIP. 
\end{alternative_coor_3dlip}
Under the $(x_0, y_0)$-invariant gait, at impact $r^2 = x_0^2 + y_0^2$. Therefore $(\alpha, \gamma, v)$ is a coordinate system for the switching manifold $\mathcal{S}$ defined in (\ref{switching_manifold_3dlip}). In the next proposition we study the Poincar\'e map of the 3D LIP in the coordinate system $(\alpha, \gamma, v)$. 
\theoremstyle{plain}
\newtheorem{3dlip_poincare}[3d_lip_invariant_gait]{Proposition}
\begin{3dlip_poincare}
Let $P: \mathcal{S}\rightarrow \mathcal{S}$ be the Poincar\'e map corresponding to the $(x_0,y_0)$-invariant gait of the 3D LIP. In the coordinate system $(\alpha, \gamma, v)$ of $\mathcal{S}$, the Poincar\'e map $P$ has the fixed point $(\alpha^*, \gamma^*, v^*)$  where $\alpha^* = \tan^{-1} (\frac{x_0}{y_0})$, $\gamma^*  =  -\omega^2 x_0 y_0$ and $v^* = \sqrt{2K_0}$.
The Jacobian of the Poincar\'e map at this fixed point is
\[
DP = \left(
\begin{array}{ccc}
0&\star & 0  \\
0&-\lambda&0\\
0&\star  & 1
\end{array}
\right ),
\]
where $\lambda$ is defined in equation (\ref{lambda_eqn}).
\label{3dlip_poincare}
\end{3dlip_poincare}
\begin{proof}
Let $P = (P_\alpha, P_\gamma, P_v)^T$. By Proposition \ref{cond_for_self_synch} under the $(x_0,y_0)$-invariant gait $\gamma^*  =  -\omega^2 x_0 y_0$ and $\alpha^* = \tan^{-1} (x_0/y_0)$. By Proposition \ref{self_sync_stability}, since at the beginning of each step $L = \gamma+\omega^2x_0 y_0$,
\begin{eqnarray*}
\lim_{L_0 \rightarrow 0} \frac{L_1}{L_0} &=& \frac{\partial P_\gamma}{\partial \gamma} \vert_{\gamma = \gamma^*}\\
&=& -\lambda
\end{eqnarray*}
where $\lambda$ is defined in equation (\ref{lambda_eqn}). 
Because the legs are massless the value of $\alpha$ at impact won't effect the dynamics of the 3D LIP biped in the next step, hence, $\frac{\partial P}{\partial \alpha} = (0,0,0)^T$.
Finally, by Proposition \ref{constant_kinetic_energy},  $v^* = \sqrt{2K_0}$, and by Proposition \ref{cond_for_self_synch} under the $(x_0, y_0)$-invariant gait $P(\alpha^*, \gamma^*, v) = (\alpha^*, \gamma^*, v)$. This proves that the last column of $DP$ is $(0,0,1)^T$. 
\end{proof}

\theoremstyle{plain}
\newtheorem{corollary_of_3dlip_poincare}[3d_lip_invariant_gait]{Corollary}
\begin{corollary_of_3dlip_poincare}
Consider a 3D LIP biped with $E_{x_0}>0$ and $E_{y_0}<0$. Let $K_0$ be the initial kinetic energy of the 3D LIP. Define the constant energy submanifold $\mathcal{S}_{K_0}$ of the switching manifold $\mathcal{S}$ as follows:
\begin{eqnarray*}
\mathcal{S}_{K_0} = \{ (\alpha, \gamma, v) \in \mathcal{S} | v = \sqrt{2K_0}\}.
\end{eqnarray*}
Under the $(x_0, y_0)$-invariant gait, the restricted Poincar\'e map $P_{K_0}: \mathcal{S}_{K_0} \rightarrow \mathcal{S}_{K_0}$ is well defined and the eigenvalues of $P_{K_0}$ are $\{-\lambda, 0\}$ with $\lambda$ defined in equation (\ref{lambda_eqn}).
\label{corollary_of_3dlip_poincare}
\end{corollary_of_3dlip_poincare}
This Corollary shows that by definition of $\lambda$, if $y_0>x_0$, after a perturbation the level of the kinetic energy of the 3D LIP might change but its motion remains self-synchronized. 
\section{The 9-DOF 3D Biped}
\label{9-DOF_3D_Biped}
In this section we generalize the definition of the discrete invariant gait of the 3D LIP to the notion of restricted discrete invariance for a 9-DOF 3D biped which is a simplified model of ATRIAS  \cite{grizzle2014atrias}.  Under the invariance assumption for the 9-DOF 3D biped, we perform a reduction and define a restricted Poincar\'e map which naturally emerges from the reduction. Finally, we define a set of controllers that can enforce the invariance. At the end, simulation results are provided to illustrate that asymptotically stable walking is achieved. 
\subsection{Configuration}
In this section we study the configuration of the system and different generalized coordinates that we might use to represent the biped. Figure \ref{schematic} shows ATRIAS \cite{grizzle2014atrias} and its simplified model, the 9-DOF 3D biped.
\begin{figure}[ht]
\begin{center}
\includegraphics[scale = .27]{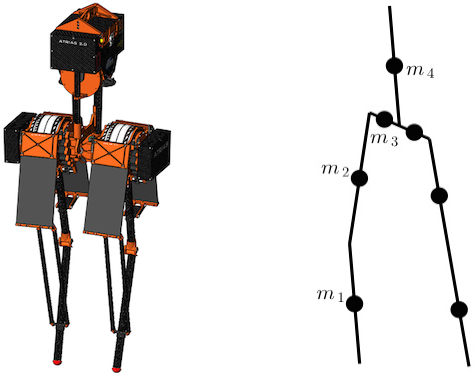}
\caption{ATRIAS vs. the 9-DOF 3D Biped}
\label{schematic}
\end{center}
\end{figure}
\begin{figure}
\begin{center}
 \begin{tikzpicture}
\linktwo(\thtwo:\Lone/2);
\jointone
\begin{scope}[shift=(\thtwo:\Lone/2),rotate=0]
	\linktwo(\ththree:\Lone/2);
	\jointtwo
          \begin{scope}[shift=(0:0),rotate=\ththree+180]
     	    	 \draw [dashed, blue] (0,0) -- (1.2\ddx,0pt);
		 \draw [<->, blue] (\Lone/4,0) arc (0:-25:\Lone/4);
		 \draw [blue] (\Lone/4,0) arc (0:-15:\Lone/4) node [below] {\small{$q_6$}}; 
          \end{scope}
	\begin{scope}[shift=(\ththree:\Lone/2), rotate=0]
		\linktwo(-\thone:\Lone/2); 
		\linktwo(85:\Lone/2); 
		\begin{scope}[shift=(85:\Lone/2),rotate=0]
		\jointone
		\end{scope}
		\draw [dashed, blue] (85:\Lone/2) -- (85:\Lone/2+\Lone/10);
		\draw [dashed, blue] (0,0) -- (180-\thone:\Lone/2+\Lone/10);
		\jointtwo
		\begin{scope}[shift=(0:0),rotate=\ththree]
			 \draw [dashed, blue] (0,0) -- (1.7\ddx,0pt);
			 \draw [<->, blue] (\Lone/2+\Lone/20,0) arc (0:10:\Lone/2+\Lone/20);
			\draw [blue] (\Lone/2+\Lone/20,0) arc (0:4:\Lone/2+\Lone/20) node [above] {\small{$q_4$}}; 
			\draw [<->,blue] (10:\Lone/2+\Lone/20)  arc (10:35:\Lone/2+\Lone/20);
			\draw [blue] (5:\Lone/2+\Lone/20) arc (5:20:\Lone/2+\Lone/20) node [above] {\small{$q_1$}};
 		\end{scope}
		\begin{scope}[shift=(-\thone:\Lone/2),rotate=0]
			\linktwo(-\thone-20:\Lone/2); 
			\draw [dashed, blue] (0,0) -- (-\thone:1.2\ddx);
			\draw [<->,blue] (-\thone:\ddx)  arc (-\thone:-\thone-20:\ddx);
			\draw [blue] (-\thone:\ddx)  arc (-\thone:-\thone-5:\ddx) node [below] {\small{$q_3$}}; 
			\jointtwo
			\begin{scope}[shift=(-\thone-20:\Lone/2),rotate=0]
			\jointone
			\robotbase
			\end{scope}
		\end{scope}
	\end{scope}
\end{scope}
\end{tikzpicture}

\vspace{.5cm}
\begin{tikzpicture}
\draw[white] (0,0)--(0,0);
\begin{scope}[shift=(3:1),rotate=0]
	\linktwo(90+\thfour:\Lone/2);
	\jointone
	\begin{scope}[shift=(90+\thfour:\Lone/2),rotate=90+\thfour]
	\linktwo(0:\Lone/2);
	\jointthree
	\end{scope}
	\begin{scope}[shift=(90+\thfour:\Lone),rotate=180+\thfour]
		\linktwo(0:\Lone/6);
		\jointthree
		\begin{scope}[shift=(0:\Lone/6),rotate=-180-\thfour]
			\linktwo(90:\Lone/2); 
			\linktwo(180-\thfour:\Lone/6);
			\jointtwo
			\node at (0,-\Lone/9.5){$H$};
		         \draw [dashed, blue] (\thfour:\Lone/6)--(\thfour:\Lone/3);
		         \draw [<->, blue] (0:\Lone/4) arc (0:\thfour:\Lone/4); 
		         \draw [blue] (0:\Lone/4) arc (0:\thfour/2:\Lone/4) node [right] {\small{$q_2$}}; 
		         \draw [dashed, blue] (180-\thfour:\Lone/6)--(180-\thfour:\Lone/3);
		          \draw [<->, blue] (180:\Lone/4) arc (180:180-\thfour:\Lone/4); 
		         \draw [blue] (180:\Lone/4) arc (180:180-\thfour/2:\Lone/4) node [left] {\small{$q_5$}}; 
		         \draw [dashed, blue] (0,0)--(0:\Lone/3);
		         \draw [dashed, blue] (0,0)--(180:\Lone/3);
			\begin{scope}[shift=(90:\Lone/2),rotate=0]
				\jointone
			\end{scope}
			
			\begin{scope}[shift=(180-\thfour:\Lone/6),rotate=180-\thfour]
				\linktwo(90:\Lone/2);
				\jointthree
				\begin{scope}[shift=(90:\Lone/2),rotate=0]
					\linktwo(90:\Lone/2);
					\jointthree
					\begin{scope}[shift=(90:\Lone/2),rotate=0]
					\jointone
					\end{scope}
				\end{scope}
					
			\end{scope}
		\end{scope}

	\end{scope}
						\robotbasetwo

\end{scope}

\end{tikzpicture}
 \end{center}
 \caption{Angle Definitions}
 \label{side_front_view_angles}
 \end{figure}
 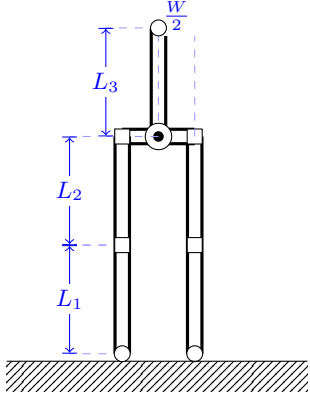
\begin{figure}
 \begin{center}
 \def\thfive{0}
\def\robotbasethree{%
    \draw (-2.7,-.1)-- (1,-.1);
    \fill[pattern=north east lines] (-2.7,-.1) rectangle (1,-.5);
}

\newcommand{\lineann}[4][0.5]{%
    \begin{scope}[rotate=#2, blue,inner sep=2pt]
        \draw[dashed, blue!40] (0,0) -- +(0,#1)
            node [coordinate] (a) {};
        \draw[dashed, blue!40] (#3,0) -- +(0,#1)
            node [coordinate] (b) {};
        \draw[|<->|] (a) -- node[fill=white] {#4} (b);
    \end{scope}
}

\begin{tikzpicture}
\begin{scope}[shift=(3:6),rotate=0]
	\linktwo(90+\thfive:\Lone/2);	
	\jointone
	\robotbase
	\begin{scope}[shift=(90+\thfive:\Lone/2),rotate=90+\thfive]
	\linktwo(0:\Lone/2);
	\jointthree
	\end{scope}
	\begin{scope}[shift=(90+\thfive:\Lone),rotate=180+\thfive]
		\linktwo(0:\Lone/6);
		\jointthree
		\begin{scope}[shift=(0:\Lone/6),rotate=-180-\thfive]
			\linktwo(90:\Lone/2); 
			\lineann[\Lone/1.8]{\thfive}{\Lone/6} {\small{$\frac{W}{2}$}}
			\linktwo(180-\thfive:\Lone/6);
			\jointtwo
			\lineann[.7]{90+\thfive}{\Lone/2} {\small{$L_3$}}
			\begin{scope}[shift=(90:\Lone/2),rotate=0]
				\jointone
			\end{scope}
			
			\begin{scope}[shift=(180-\thfive:\Lone/6),rotate=180-\thfive]
				\linktwo(90:\Lone/2);
				\lineann[-0.7]{90+\thfive}{\Lone/2} {\small{$L_2$}}
				\jointthree
				\begin{scope}[shift=(90:\Lone/2),rotate=0]
					\lineann[-0.7]{90+\thfive}{\Lone/2} {\small{$L_1$}}
					\linktwo(90:\Lone/2);
					\jointthree
					\begin{scope}[shift=(90:\Lone/2),rotate=0]
					\jointone
					\end{scope}
				\end{scope}
			\end{scope}
		\end{scope}
	\end{scope}
\end{scope}

\end{tikzpicture}
 \end{center}
 \caption{Biped Dimensions}
 \end{figure}
\subsubsection{Generalized Coordinates}
As shown in Figures \ref{side_front_view_angles}, a set of generalized coordinates that we might use to describe the biped's configuration is $q = (\theta_y, \theta_r, \theta_p, q_1, q_2, q_3, q_4, q_5, q_6)$. The Euler angles $(\theta_y, \theta_r, \theta_p)$, are the yaw, roll and pitch angle of the torso which describe the orientation of torso with respect to the inertial frame. As shown in Figure \ref{side_front_view_angles}, $(q_1, q_2, q_3, q_4, q_5, q_6)$ are relative angles with respect to the torso and they describe the configuration of the stance and swing legs in a coordinate system attached to the torso. For later reference, we define $\hat{q} = (\theta_r, \theta_p, q_1, q_2, q_3, q_4, q_5, q_6)$. 

\subsubsection{Alternative Coordinates; Quasi-Velocities}

Let $W$ denote the inertial (world) frame. Define the coordinate system $I$ to be parallel to $W$ and centered at the support point. Also, let the rotating coordinate system $Y$ be centered at the support point and rotated by $\theta_y$, where $\theta_y$ is the yaw angle of the torso with respect to I. 

Let $\boldsymbol{r}_H$ and $\boldsymbol{r}_F$ denote the position vector of the hip (see Figure \ref{side_front_view_angles}) and swing leg end in  W, respectively. Define $\boldsymbol{r}_{FH} = \boldsymbol{r}_H - \boldsymbol{r}_F$. Let $(x_{I}, y_{I}, z_{I})$ and $(x_{{FH}_I}, y_{{FH}_I}, z_{{FH}_I})$ be the coordinates of  $\boldsymbol{r}_H$ and $\boldsymbol{r}_{FH} $  in I and let $(x, y, z)$, $(x_{FH}, y_{FH}, z_{FH})$, $(v_x, v_y, v_z)$ be the coordinates of $\boldsymbol{r}_H$, $\boldsymbol{r}_{FH}$ and $\dot{\boldsymbol{r}}_H$ in Y. 
\theoremstyle{proposition}
\newtheorem{hip_vel_eqns}[3d_lip_invariant_gait]{Lemma}
\begin{hip_vel_eqns}
We have $x v_x + y v_y + z v_z = x_I\dot{x}_I + y_I \dot{y}_I + z_I \dot{z}_I.$
\label{hip_vel_eqns}
\end{hip_vel_eqns}
In the next proposition we introduce two alternative sets of generalized coordinates for the 9-DOF 3D biped, which will be used later. 
\theoremstyle{plain}
\newtheorem{alternative_coordinates}[3d_lip_invariant_gait]{Proposition}
\begin{alternative_coordinates}
If 
\begin{eqnarray*}
\xi_1& =& (\theta_y, \theta_r, \theta_p, x, y, z, x_{FH}, y_{FH}, z_{FH})\\
\zeta_1 &=& (\dot{\theta}_y, \dot{\theta}_r, \dot{\theta}_p, v_x, v_y, v_z, \dot{x}_{FH}, \dot{y}_{FH}, \dot{z}_{FH}),
\end{eqnarray*}
then $(\xi_1, \zeta_1)$ is a coordinate system for $\mathcal{TQ}$, the tangent bundle of the configuration space $\mathcal{Q}$. 
Assuming $y\ne 0$, let
\begin{eqnarray*}
\begin{array}{llllll}
\alpha &=& \tan^{-1} (\frac{x}{y}), &r &=& \sqrt{x^2+y^2}\\
\gamma &=& v_x v_y, &v& =& \sqrt{v_x^2+v_y^2},
\end{array}
\end{eqnarray*}
with $\alpha \in [-\pi/2, \pi/2]$, and define
\begin{eqnarray*}
\xi_2 &=& (\theta_y, \theta_r, \theta_p, r, \alpha, z, x_{FH}, y_{FH}, z_{FH})\\
\zeta_2 &=& (\dot{\theta}_y, \dot{\theta}_r, \dot{\theta}_p, v, \gamma, v_z, \dot{x}_{FH}, \dot{y}_{FH}, \dot{z}_{FH})\,.
\end{eqnarray*}
Then $(\xi_2, \zeta_2)$ is a coordinate system for $\mathcal{TQ}$. Moreover, if $\hat{\xi}_2 = (\theta_r, \theta_p, r, \alpha, z, x_{FH}, y_{FH}, z_{FH})$, then $(\hat{\xi}_2, \zeta_2)$ is a function of $(\hat{q}, \dot{q})$. Furthermore, if $\mathcal{S} = \{(q,\dot{q}) | z_{F}(q) = 0\}$, then,
\begin{eqnarray*}
\xi_\mathcal{S} &=& (\theta_y, \theta_r, \theta_p, r, \alpha, z, x_{FH}, y_{FH})\\
\zeta_\mathcal{S} &=& (\dot{\theta}_y, \dot{\theta}_r, \dot{\theta}_p, v, \gamma, v_z, \dot{x}_{FH}, \dot{y}_{FH}, \dot{z}_{FH}),
\end{eqnarray*}
defines a coordinate system on $\mathcal{S}$.  

\label{alternative_coordinates}
\end{alternative_coordinates}
Note that $\zeta_1$ and $\zeta_2$ are not the derivatives of $\xi_1$ and $\xi_2$. They are called \textit{quasi-velocities} \cite{bloch2009quasivelocities}.

\subsection{Equations of Motion}
Walking is modeled  as having continuous and discrete phases. The continuous phase is governed by the Euler-Lagrange equations. The equations of motion in the continuous phase are of the form 
\begin{equation}
D(q)\ddot{q}+H(q,\dot{q})=Bu,
\label{euler_lagrange_eqns}
\end{equation}
 where $u_{6\times 1}=[u_S,u_F]^T$ with $u_S$ and $u_F$ denoting the controllers in the stance and swing legs respectively. Also, $B=[0_{3\times 6};I_{6\times 6}]$. In the discrete phase the velocities are transformed instantly due to the change of support point (from left to right or right to left). Using methodology of \cite{hurmuzlu1994rigid} which is illustrated for legged robots in \cite{westervelt2007feedback} we obtain the impulse matrix $\Delta_q (q)$ which maps the velocities $\dot{q}^-$ right before the impact to the velocities immediately after impact, $\dot{q}^+$. In order to use the same set of equations when either the left or the right leg is the stance leg, we implement a matrix $R$ to swap the roles of stance and swing legs. Therefore, setting $x=(q,\dot{q})$, if we rewrite equation (\ref{euler_lagrange_eqns}) and the impulse map in terms of $x$ and $\dot{x}$, we obtain 

\begin{equation*}
\Sigma=
\left \{
\begin{array}{lllc}
\dot{x}&=&f(x)+g(x)u&x^- \notin \mathcal{S}\\
x^+&=&\Delta(x^-) & x^-\in \mathcal{S},
\end{array}
\right.
\end{equation*}
where $\Delta(q,\dot{q}^-)= (Rq,R\Delta_q(q)\dot{q}^-)$ and $\mathcal{S}$ is the \textit{switching manifold}, which is defined as 
\begin{equation*}
\mathcal{S} = \{(q,\dot{q}) | z_{F}(q) = 0\},
\label{9dof_impact_surface}
\end{equation*}
where $z_{F}$ is the height of the swing leg end in W. 

\theoremstyle{proposition}
\newtheorem{continuous_indep_of_thy}[3d_lip_invariant_gait]{Lemma}
\begin{continuous_indep_of_thy}
The kinetic and potential energies of the 9-DOF 3D biped are independent of $\theta_y$ \cite{spong2005controlled}. As a result, under the assumption that the controllers are independent of $\theta_y$, if $\theta_{y_0}$ is the initial value of $\theta_y$ right after the impact, the evolution of $(\hat{q}, \dot{q})$ and $\theta_y - \theta_{y_0}$ are independent of $\theta_{y_0}$.
\label{continuous_indep_of_thy}
\end{continuous_indep_of_thy}

\theoremstyle{proposition}
\newtheorem{impact_indep_of_thy}[3d_lip_invariant_gait]{Lemma}
\begin{impact_indep_of_thy}
The impact map is independent of $\theta_y$ in the sense that $\dot{q}^+ = \Delta_{q}(\hat{q}) \dot{q}^-$ \cite{spong2005controlled}.
\label{impact_indep_of_thy}
\end{impact_indep_of_thy}

\subsection{Restricted Invariance and Reduction}

\label{Reduction}
In this section we introduce a generalization of the discrete invariant gait of the 3D LIP to the 9-DOF 3D biped. This gait enables us to reduce the system dimension to four dimensions and to describe the dynamics in a set of variables
which are particularly amenable to analysis. 

\theoremstyle{definition}
\newtheorem{M_invariant_step}[3d_lip_invariant_gait]{Definition}
\begin{M_invariant_step}
Let $\mathcal{M}$ be a submanifold of the switching surface $\mathcal{S}$ of the 9-DOF 3D biped. We say that the biped completes an $\mathcal{M}$-invariant step if the solutions starting in $\mathcal{M}$ end in $\mathcal{M}$. 
\end{M_invariant_step}

\theoremstyle{plain}
\newtheorem{symmetric_gait}[3d_lip_invariant_gait]{Proposition}
\begin{symmetric_gait}
Let $\theta_p^d>0, x_0>0, y_0>0$ and $q_k^d>0$, and define the quadruple $\mathcal{P} = (\theta_p^d, x_0, y_0, q_k^d)$.  Suppose that the 9-DOF 3D biped completes a step such that at the time of impact, 
\begin{enumerate}[(i)]
\item $\theta_{p} = \theta_p^d$, $\theta_{r} = 0$, $q_{3} = q_k^d$, $\dot{\theta}_{p} = 0$, $\dot{\theta}_{r} = 0$, $\dot{q}_{3} = 0$
\item $x_{{FH}} = -x_0$, $y_{{FH}} = -y_0$, $q_{6} = q_k^d$, $\dot{x}_{{FH}} = 0$, $\dot{y}_{{FH}} = 0$,  $\dot{q}_{6} = 0.$
\end{enumerate}
 Then the biped has completed an $\mathcal{M}_\mathcal{P}$-invariant step, where, $\mathcal{M}_\mathcal{P}$ is a submanifold of $\mathcal{S}$ which is the image of the local embedding $f_\mathcal{P}:\mathcal{S}\rightarrow \mathcal{S}$ defined as $f_\mathcal{P}(\xi_\mathcal{S}, \zeta_\mathcal{S}) = (\xi_\mathcal{P}, \zeta_\mathcal{P})$, where
\begin{eqnarray*}
\xi_\mathcal{P} &=& (\theta_y, 0, \theta_p^d, r_0, \alpha,z_0, x_0, y_0) \\
\zeta_\mathcal{P} &=& (\dot{\theta}_y, 0, 0, v, \gamma, v_z, 0, 0, 0), 
\end{eqnarray*}
\label{symmetric_gait}
for constants $r_0$ and $z_0$ which are functions of $\mathcal{P}$. Moreover, $\mathcal{M}_\mathcal{P}$ is 5-dimensional and,
\begin{eqnarray*}
\xi_r = (\theta_y, \alpha), \zeta_r  =  (\gamma, \dot{\theta}_y, v), 
\end{eqnarray*}
defines a coordinate system on $\mathcal{M}_\mathcal{P}$. In the closed-loop system the evolution of $(q, \dot{q})$ in the next step is uniquely determined by the value of $(\xi_r, \zeta_r)$ at impact. 

\end{symmetric_gait}
The goal of the controllers is then that of rendering $\mathcal{M}_\mathcal{P}$ invariant. We will discuss this in more details in Section \ref{Controllers}. 
\begin{proof}
\textit{Step 1.} We first show that, by assumptions $(i)$ and $(ii)$ we have $x^2 + y^2  =  x_0^2+y_0^2$ and $z = z_0$, where
\begin{eqnarray*}
z_0 & = & \sqrt{r_1^2-x_0^2-y_0^2},\\
r_1^2 &=& L_1^2 + L_2^2 +W^2/4+2 L_1 L_2 \cos(q_k^d).
\end{eqnarray*}
From the kinematic equations of the biped, 
\begin{eqnarray}
\label{rH2}
\Vert \boldsymbol{r}_H\Vert^2 &=& L_1^2 + L_2^2 +\frac{W^2}{4}+2 L_1 L_2 \cos(q_3)\label{rHsquare},\\
\label{rFH2}
\Vert \boldsymbol{r}_{FH}\Vert^2 &=& L_1^2 + L_2^2 +\frac{W^2}{4}+2 L_1 L_2 \cos(q_6).
\end{eqnarray}
At impact, $x_{FH} = -x_0$, $y_{FH} = -y_0$ and $q_6 = q_k^d$. Therefore, looking at equation (\ref{rFH2}), we have $x_0^2+y_0^2+z_{FH}^2 = r_1^2$.  Thus, by definition of $z_0$, at impact $z_{FH} = z_0$. Consequently, since $z_{FH} = z - z_{F}$ and at impact $z_{F} = 0$, we have $z= z_0$. Also, since at impact $q_6 = q_3 = q_k^d$, from equations (\ref{rH2}) and (\ref{rFH2}) at impact $\Vert \boldsymbol{r}_H\Vert = \Vert \boldsymbol{r}_{FH}\Vert$. Hence, $x^2+y^2+z^2 = x_0^2+y_0^2+z_0^2$. Since we just showed that at impact $z = z_0$, we have $x^2+y^2 = x_0^2+y_0^2$.  
\label{geom_eqns}

\textit{Step 2.} By Propositions \ref{alternative_coordinates} and Step 1, at impact,
\begin{eqnarray*}
\xi_2 &=& (\theta_y, 0, \theta_p^d, r_0, \alpha, z_0, x_0, y_0, z_0) \\
\zeta_2 &=& (\dot{\theta}_y, 0, 0, v,\gamma, v_z, 0, 0, 0), 
\end{eqnarray*}
where $r_0 = \sqrt{x_0^2+y_0^2}$. By equation (\ref{rHsquare}), Lemma \ref{hip_vel_eqns} and the assumption that at impact $\dot{q}_3 = 0$, we have $v_z = -\frac{1}{z_0} (xv_x+yv_y)$. Therefore, $v_z$ is a smooth function of $\alpha$, $ \gamma$ and $v$. So, the reduced coordinates of $\mathcal{M}_\mathcal{P}$ at impact are $\xi_r = (\theta_y, \alpha), \zeta_r  =  (\gamma, \dot{\theta}_y, v)$. Thus the evolution of $(q, \dot{q})$ in the next step is uniquely determined by the value of $(\xi_r, \zeta_r)$ at impact. 
\end{proof}

\theoremstyle{plain}
\newtheorem{dependence_on_four_var}[3d_lip_invariant_gait]{Corollary}
\begin{dependence_on_four_var}
If the biped performs an $\mathcal{M}_\mathcal{P}$-invariant step and the controllers are independent of $\theta_y$, then the evolution of $(\alpha, \gamma, \dot{\theta}_y, v)$ and $\theta_y-\theta_{y_0}$ in the next step is uniquely determined by the value of $(\alpha, \gamma, \dot{\theta}_y, v)$ at impact. 
\begin{proof}
Since, as stated in Proposition \ref{alternative_coordinates}, $(\alpha, \gamma, \dot{\theta}_y, v)$ is only a function of $(\hat{q},\dot{q})$, the proof follows from Lemma \ref{continuous_indep_of_thy}, Lemma \ref{impact_indep_of_thy} and Proposition \ref{symmetric_gait}. 
\end{proof}
\label{dependence_on_four_var}
\end{dependence_on_four_var}

\theoremstyle{plain}
\newtheorem{embedded_submanifold_4_dim}[3d_lip_invariant_gait]{Corollary}
\begin{embedded_submanifold_4_dim}
Corresponding to an $\mathcal{M}_\mathcal{P}$-invariant step there exists a 4-dimensional invariant embedded submanifold $\hat{\mathcal{M}}_{\mathcal{P}}$ of the switching manifold $\mathcal{S}$ such that $(\alpha, \gamma, \dot{\theta}_y, v)$ is a coordinate system of $\hat{\mathcal{M}}_{\mathcal{P}}$.
\label{embedded_submanifold_4_dim}
\end{embedded_submanifold_4_dim}

\theoremstyle{definition}
\newtheorem{restricted_poincare_map}[3d_lip_invariant_gait]{Definition}
\begin{restricted_poincare_map}
Assuming that ${\mathcal{M}}_{\mathcal{P}}$ is invariant, i.e. solutions starting from ${\mathcal{M}}_{\mathcal{P}}$ end in ${\mathcal{M}}_{\mathcal{P}}$, we can define the $\hat{\mathcal{M}}_{\mathcal{P}}$-restricted Poincar\'e  map, $\hat{P}:\hat{\mathcal{M}}_{\mathcal{P}}\rightarrow \hat{\mathcal{M}}_{\mathcal{P}}$, as the function that maps the value of $(\alpha, \gamma, \dot{\theta}_y, v)$ at the end of the current step to its value at the end of the next step. 
\label{restricted_poincare_map}
\end{restricted_poincare_map}

\theoremstyle{definition}
\newtheorem{alternating_coor_sys}[3d_lip_invariant_gait]{Remark}

\begin{alternating_coor_sys}
For simplicity, we assume that the coordinate systems $I$ and $Y$ are right handed when the right leg is the stance leg and they are left handed when the left leg is the stance leg. This choice allows us to analyze the periodic motion and its stability in one step rather than two steps. 
\end{alternating_coor_sys}

\theoremstyle{plain}
\newtheorem{periodicity_thy}[3d_lip_invariant_gait]{Proposition}
\begin{periodicity_thy}
If the submanifold ${\mathcal{M}}_{\mathcal{P}}$ is invariant and the controllers are independent of $\theta_y$, then the restricted Poincar\'e map has a fixed point $x^*_r = (\alpha^*, \gamma^*, \dot{\theta}_y^*, v^*)$ if and only if the hybrid system has a periodic solution for $(\hat{q}, \dot{q})$ which passes through $x^*_r$. This periodic orbit is asymptotically stable if and only if $x^*_r$ is an asymptotically stable fixed point of $\hat{P}$. Moreover, if the fixed point $x^*_r$ exists and is asymptotically stable, then $\theta_y$ is 2-periodic and is neutrally stable. 
\label{periodicity_thy}
\end{periodicity_thy}
\begin{proof}(\textit{Sketch})
The equivalence of periodicity and asymptotic stability of $(\hat{q},\dot{q})$ with $\hat{P}$ having an asymptotically stable fixed point follows from Corollary \ref{dependence_on_four_var}. Also,  because by Corollary \ref{dependence_on_four_var}, the evolution of $\theta_y$ depends on $\theta_{y_0}$,  we can show that $\theta_y$ is 2-periodic and is neutrally stable (it is not asymptotically stable because $\theta_{y_0}$ might change due to a perturbation even though $\theta_y-\theta_{y_0}$ is stable). 
\end{proof}

By this proposition, if we find the controllers that render $\mathcal{M}_\mathcal{P}$ invariant, then the periodicity and stability of the motion of the biped can be checked by just looking at the variables $(\alpha, \gamma, \dot{\theta}_y, v)$ at impact. In the next section we discuss such controllers.

\subsection{Controllers}
\label{Controllers}
To achieve an $\mathcal{M}_\mathcal{P}$-invariant gait, we find the controllers $u_S$ and $u_F$ such that the conditions in Proposition \ref{symmetric_gait} hold. To this end, we define two goals for our control algorithm: posture control and foot placement. The goal of the posture control is to make $y_1$ zero and the foot placement algorithm drives $y_2$ to zero, where
\[
y_1=
\left(
\begin{array}{c}
 \theta_r  \\
 \theta_p-\theta_p^d\\
 q_3
\end{array}
\right),
y_2=
\left(
\begin{array}{c}
 x_{FH}-x_0  \\
 y_{FH}-y_0\\
 q_6-q_6^d
\end{array}
\right).
\]
To let the swing foot clear the ground, in the first half of the step $q_6^d$ is nonzero and in the second half it is set to zero. For posture control, we only use $u_S$. If $u_F$ is involved in the posture control then the flight leg would swing in a direction opposite to the torque it applies to the torso. This will disrupt foot placement. Thus, $u_F$ is only used to accomplish the foot placement. As a comparison to the 3D LIP, this foot placement algorithm aims to achieve an $(x_0, y_0)$-invariant gait. More details of the controllers will appear in a future paper.
\subsection{Simulation Results}
\label{Simulation_Results}
Under the above controllers, the biped was able to start walking from zero velocity and exhibit a stable gait. The animations can be found in \cite{razavi2014animation3D}.   Figure \ref{graph_of_vars} shows how the sequence $(\alpha_n, \gamma_n, \dot{\theta}_{y_n}, v_n)$ converges to a fixed point. The dominant eigenvalue of the restricted Poincar\'e map was found to have an absolute value of 0.96.  
\begin{figure}
\begin{center}
\includegraphics[scale = .51]{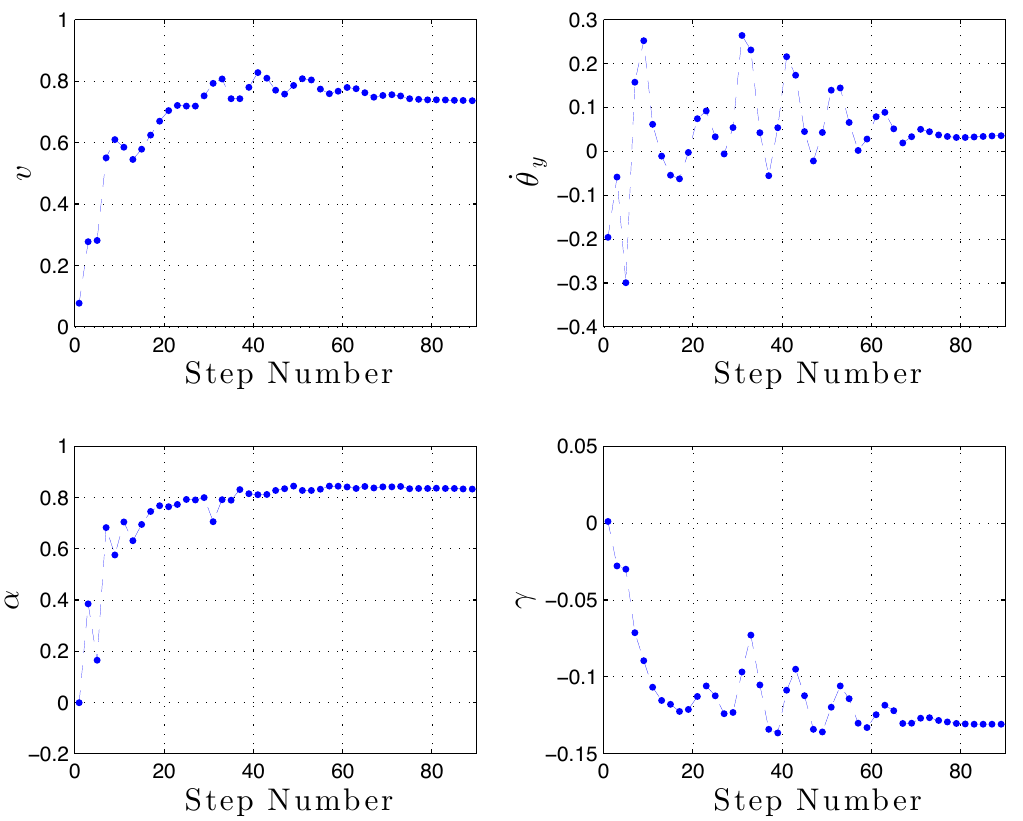}
\end{center}
\caption{$(\alpha, \gamma, \dot{\theta}_y, v)$ vs. Step Number. Only the states when the left leg is the stance leg are shown.}
\label{graph_of_vars}
\end{figure}
\section{Conclusion}
In this paper, we introduced a discrete invariant gait for the 3D LIP and we proved that under this gait the 3D LIP is self-synchronized and neutrally stable with respect to the kinetic energy. Then we generalized the definition of the discrete invariant gait to the notion of $\mathcal{M}_\mathcal{P}$-invariant gait of a 9-DOF 3D biped. After performing a reduction based on this invariant gait we showed that the periodicity and asymptotic stability of the motion of the 9-DOF 3D biped can be determined by studying a restricted Poincar\'e map in four variables. Finally, we provided an example of a controller that may satisfy the conditions of the $\mathcal{M}_\mathcal{P}$-invariant gait. By applying this controller to the 9-DOF 3D biped we showed numerically that it performs stable walking. The $\mathcal{M}_\mathcal{P}$-invariant gait introduced here is not limited to the particular 9-DOF 3D biped which was studied here. For example, we tested this gait on a half scale biped model of the 9-DOF 3D biped we studied here and were able to achieve stable walking. 
\vspace{.1cm}

{\bf Acknowledgements:} We gratefully acknowledge partial support by NSF INSPIRE grant 1343720. 
\bibliography{/Users/Hamed/Research/PhD_research/References_Bib_file/mybiped}{}

\bibliographystyle{plain}

\end{document}